# MultiCNKG: Integrating Cognitive Neuroscience, Gene, and Disease Knowledge Graphs Using Large Language Models

**Ali Sarabadani\* [1], Kheirolah Rahsepar Fard [2]**

1- Ph.D. in IT Engineering , Department of Computer Engineering and Information Technology, University of Qom, Qom, Iran.
a.sarabadani@stu.qom.ac.ir.

2- Department of Computer Engineering and Information Technology, University of Qom, Qom, Iran. rahsepar@qom.ac.ir

 \* a.sarabadani@stu.qom.ac.ir

**Abstract**

The advent of large language models (LLMs) has revolutionized the integration of knowledge graphs (KGs) in biomedical and cognitive sciences, overcoming limitations in traditional machine learning methods for capturing intricate semantic links among genes, diseases, and cognitive processes. We introduce MultiCNKG, an innovative framework that merges three key knowledge sources: the Cognitive Neuroscience Knowledge Graph (CNKG) with 2.9K nodes and 4.3K edges across 9 node types and 20 edge types; Gene Ontology (GO) featuring 43K nodes and 75K edges in 3 node types and 4 edge types; and Disease Ontology (DO) comprising 11.2K nodes and 8.8K edges with 1 node type and 2 edge types. Leveraging LLMs like GPT-4, we conduct entity alignment, semantic similarity computation, and graph augmentation to create a cohesive KG that interconnects genetic mechanisms, neurological disorders, and cognitive functions. The resulting MultiCNKG encompasses 6.9K nodes across 5 types (e.g., Genes, Diseases, Cognitive Processes) and 11.3K edges spanning 7 types (e.g., Causes, Associated with, Regulates), facilitating a multi-layered view from molecular to behavioral domains. Assessments using metrics such as precision (85.20%), recall (87.30%), coverage (92.18%), graph consistency (82.50%), novelty detection (40.28%), and expert validation (89.50%) affirm its robustness and coherence. Link prediction evaluations with models like TransE (MR: 391, MRR: 0.411) and RotatE (MR: 263, MRR: 0.395) show competitive performance against benchmarks like FB15k-237 and WN18RR. This KG advances applications in personalized medicine, cognitive disorder diagnostics, and hypothesis formulation in cognitive neuroscience.

## 1- Introduction

Knowledge graphs (KGs) are widely utilized across various fields and applications, addressing numerous real-world scenarios [1]. Many of these graphs, such as citation graphs [2] and product graphs [3], feature nodes linked to textual attributes, which contribute to text-related occurrences. For instance, in the Ogbn-products dataset [2], each node symbolizes a product, with its textual description serving as a node attribute. These graphs have found extensive use across diverse domains, including social network analysis [4], information retrieval [5], and a broad spectrum of natural language processing (NLP) tasks. Among the most significant KGs are biomedical graphs, which excel in uncovering complex interconnections among biological systems, diseases, and genes. Given the prevalence of text-attributed graphs (TAGs), our objective is to represent this graph using three knowledge sources: the Cognitive Neuroscience Knowledge Graph (CNKG), Disease Ontology (DO), and Gene Ontology (GO). Graph Neural Networks (GNNs) [6] have become a standard approach for managing graph-structured data, employing a message-passing mechanism to efficiently capture graph structures. Traditional methods often rely on asymmetric shallow embeddings like Bag-of-Words [7] and Word2Vec [8] to encode textual data, though these approaches face limitations. Recent research highlights that these low-content embeddings struggle with capturing polysemous words [9] and lack semantic depth [10], potentially reducing effectiveness in downstream tasks. In contrast, large language models (LLMs) offer extensive context-aware knowledge and enhanced semantic understanding through pre-training on vast text corpora [8]. This pre-trained knowledge has sparked a revolution in downstream NLP tasks [11]. Models like ChatGPT and GPT-4 [12], boasting hundreds of billions of parameters, demonstrate exceptional performance across various text-related domains. This raises an important question: Can LLMs compensate for the contextual and semantic shortcomings of traditional GNN pipelines? LLM-based approaches have increasingly supplanted conventional methods, leveraging pre-trained knowledge and excelling in tasks with implicit graph structures, such as recommendations, ranking, and multi-hop reasoning, where they are used for final predictions. This success reinforces the capability of LLMs to handle complex challenges.

The evolution of transformer-based pre-trained language models has transformed NLP and related tasks. These architectures encompass encoder-like models (e.g., BERT [13]), decoder models (e.g., GPT [14]), and encoder-decoder models (e.g., T5 [15]). Transformer-based language models, including OpenAI's GPT-4 [12], Google's Gemini [16], PaLM [17], Microsoft's Phi 3 [18], and Meta LLaMA [19], are collectively termed LLMs. These models are trained on large-scale transformers with billions of parameters, enabling agent-like capabilities such as perception, reasoning, planning, and action [20].

The primary contribution of this study is the development of MultiCNKG-LLM, a novel method designed to integrate data from three knowledge bases: CNKG, DO, and GO. MultiCNKG-LLM not only retains critical relationships from the original datasets but also reveals new, previously undiscovered connections. This innovative approach marks a significant advancement in knowledge graph applications.

## 2- Related works

Knowledge graphs (KGs) have emerged as a robust tool for presenting and integrating complex biomedical data, facilitating efficient data synthesis and enabling researchers to identify novel relationships among entities like diseases, genes, and drugs. Notable biomedical KGs include Bio2RDF [21] and CTKG by Chen et al. [22].

The field of biomedical knowledge processing has also advanced significantly with the adoption of large language models (LLMs), which enhance the interpretation of medical literature, clinical notes, and diagnostic support. Prominent LLMs in healthcare include DeepMind's MedIC, Microsoft's BioGPT [23], TrialGPT [24], BioBart [25], and BioMistral [26]. These models have had a profound impact, particularly in drug discovery, patient data analysis, and clinical decision-making, necessitating the processing of vast amounts of unstructured data.

Recent studies support the integration of LLMs to develop DIAMOND-KG, a knowledge graph that captures drug indications alongside their clinical context [27]. This research demonstrated that a sophisticated text extraction strategy substantially improves data quality, with 71% of drug indications linked to at least one relevant clinical reference, underscoring the value of reference data in enhancing KG utility.

The integration of LLMs with KGs is rapidly expanding in medicine, neuroscience, and biomedicine, offering new avenues for knowledge acquisition and healthcare applications. Researchers have created advanced platforms, such as that by Mann et al. [28], which combine existing medical knowledge sources into a comprehensive KG to support practical treatment identification based on known symptoms or diseases.

Fast LLM techniques have gained traction in clinical concept extraction [29], improving the accuracy of extracting medical concepts from unstructured data. However, challenges persist, particularly in distinguishing drug codes with relevant medical contexts. Despite progress, many automated methods still struggle to clarify these critical details. Recent investigations highlight the significant potential of LLMs for graph-based tasks, including relationship extraction, KG completion, and querying [30]. For instance, Khorashadizadeh et al. [31] conducted a qualitative analysis using decoders like ChatGPT and Bard for KG integrity and biomedical queries, finding that while LLMs like ChatGPT show promise in automatic KG extraction, computational costs remain high.

With the rise of generative AI, KG evaluation has become a key research area. Studies indicate that LLMs can automatically assess KG health models [32], offering a potential replacement for human verification, thus improving efficiency and scalability. Additionally, LLMs have been employed to evaluate class member relationships in public KGs like Wikidata and Calligraph, aiding knowledge engineers in KG enhancement [33].

In biomedical applications, various techniques have been employed to optimize KG operations through semantic similarity searches using embedded vectors. Methods like embedded search followed by nearby data retrieval, query classification to organize filters and summarize LLMs [34][35], and advanced approaches such as neighborhood descriptors, query rewriting, logical query structure breakdown, and fine-tuning model weights with KG training data [36][37][38][39] have been utilized

LLMs have demonstrated remarkable performance improvements in medical tasks, including disease diagnosis and prediction using electronic medical records (EMR). For example, Jiang et al. [40] developed a patient-specific KG using a bidirectional attention-enhanced graph neural network (BAT GNN) to enhance medical decision-making. RAM-EHR converts multiple knowledge sources into text format via co-training, improving information extraction for better medical predictions [41]. DR.KNOWS boosts diagnostic accuracy and interpretation by integrating KGs with clinical diagnostic logic models based on the Unified Medical Language System (UMLS) [42]. Furthermore, REALM enhances efficiency in complex medical cases by integrating clinical records and multivariate time series using Retrieval-Augmented Generation (RAG) technology.

## 3- Methodology

The proposed methodology for **MultiCNKG** consists of four major stages:

1. **Data Collection and Preprocessing**
2. **Knowledge Graph Representation**
3. **Graph Alignment and Integration**
4. **Graph Expansion using Large Language Models (LLMs)**

Each stage is described in detail below.

### 3.1 Data Collection and Preprocessing

Three knowledge sources were selected:

- **Cognitive Neuroscience Knowledge Graph (CNKG):** Represents concepts, tasks, and processes in cognitive neuroscience, such as memory, attention, decision-making, and learning, along with their neural correlates.
- **Gene Ontology (GO):** Provides structured annotations of genetic and molecular functions.
- **Disease Ontology (DO):** Encodes diseases and disorders, including neurological and psychiatric conditions.

**Table 1 – CNKG , DO and GO**

| Name | # Node | # Node Types | # Edges | # Edge Types |
|---|---|---|---|---|
| Cognitive Neuroscience Knowledge Graph (CNKG) | 2.9 K | 9 | 4.3 K | 20 |
| Disease Ontology | 11.2 K | 1 | 8.8 K | 2 |
| Gene Ontology | 43 K | 3 | 75 K | 4 |

The datasets were preprocessed using LLM-based pipelines (GPT-4, BioGPT). Preprocessing steps included:

1. **Tokenization:** Raw text descriptions of entities were tokenized into sequences:

$$T = \{t_1, t_2, \dots, t_n\}$$

where each $ti$ represents a token.

2. **Normalization:** To ensure consistency, entities with multiple forms (e.g., *Alzheimer's disease* and *AD*) were mapped to canonical forms using embedding similarity:

$$\sim(e_i, e_j) = \frac{v_i \cdot v_j}{\|v_i\|\|v_j\|}$$

where $vi, vj$ are vector embeddings generated by LLMs.

2. **Noise Removal:** Duplicate or irrelevant nodes were removed using threshold-based filtering on similarity scores.

### 3.2 Knowledge Graph Representation

The integrated KG is defined as a directed multigraph:

$$G = (V, E, R)$$

where:

- V is the set of nodes (genes, diseases, cognitive neuroscience concepts),
- E ⊆ V×R×V is the set of edges,
- R is the set of relations (e.g., *causes*, *associated_with*, *impairs*).

Each edge is represented as a **knowledge triple**:

$$(h, r, t) \text{ with } h, t \in V, r \in R$$

For example:

(APOE4, causes, Alzheimer's disease)

(Alzheimer's disease, impairs, Memory)

The graph structure can also be encoded in an **adjacency matrix**:

$$A_{ij} = \begin{cases} 1, & \text{if there exists a relation between } v_i \text{ and } v_j \\ 0, & \text{otherwise} \end{cases}$$

### 3.3 Graph Alignment and Integration

To merge CNKG, GO, and DO, we align nodes and edges that represent the same or semantically related entities.

1. **Node Alignment:** For nodes $vi \in V_{CNKG}$, $vj \in V_{DO}$, and $vk \in V_{GO}$, the final unified set is:

$$V_f = V_{CNKG} \cup V_{DO} \cup V_{GO}$$

Similar nodes are merged if their embedding similarity satisfies:

$$\sim(v_i, v_j) \geq \tau$$

where τ\tau is a predefined threshold.

2. **Edge Alignment:** Edges that represent equivalent relations (e.g., *causes* vs. *induces*) are unified by LLM-based semantic comparison. The final edge set is:

$$E_f = E_{CNKG} \cup E_{DO} \cup E_{GO}$$

3. **Adjacency Update:** The adjacency matrix is updated as:

$$A_f = A_{CNKG} + A_{DO} + A_{GO} + \Delta A$$

where $\Delta A$ represents newly discovered relations.

### 3.4 Graph Expansion with LLMs

The most critical step is leveraging LLMs to discover **new relationships** that are not explicitly present in the original graphs.

1. **Relation Prediction:** For two nodes $h, t \in V$, the probability of a new relation $r\,new$ is estimated by LLM:

$$P(r_{new} \mid h, t) = f_{LLM}(h, t)$$

A relation is added if:

$$P(r_{new} \mid h, t) \geq \tau$$

2. **Probabilistic Similarity Model:** Alternatively, similarity-based relation discovery is modeled as:

$$P(r_{new}) = 1 - \exp\left(-\frac{\|v_h - v_t\|^2}{2\sigma^2}\right)$$

Where $vh, vt$ are embeddings of nodes $h, t,$ and $\sigma$ controls distribution spread.

3. **Iterative Expansion:** The KG evolves iteratively:

$$G^{(t+1)} = G^{(t)} + \Delta G^{(t)}$$

Where $\Delta G^{(t)} = (\Delta V^{(t)}, \Delta E^{(t)})$ are new entities and edges discovered at iteration tt.

4. **Validation:** Newly discovered relations are assigned a confidence score and validated either automatically via external databases or manually by domain experts. If feedback decreases confidence below threshold:

$$P(r_{new}) \leftarrow P(r_{new}) - \delta P$$

the relation is removed in the next iteration.

### 3.5 Final Integrated Graph

The final **MultiCNKG** integrates three layers:

1. **Genetic Layer (GO):** Molecular and functional gene information.
2. **Disease Layer (DO):** Neurological and psychiatric disorders.
3. **Cognitive Neuroscience Layer (CNKG):** Memory, attention, learning, and decision-making constructs.

The integration allows tracing **pathways from genes → diseases →cognitive functions**, offering a holistic perspective for cognitive neuroscience research.

## 4. Simulation and Results

LLM was used to build the knowledge graph and both types of nodes and edges for MultiCNKG. This linguistic model extracted entities and edges from raw texts related to cognitive neuroscience, gene ontology, and disease ontology articles. This model identified 5 different node types (refer to Table 2) and 7 different edge types (refer to Table 3). The strength of this model lies in its training to recognize diverse biomedical and cognitive neuroscience terms and their contexts, allowing for the distinction between various types of nodes and relationships.

**Table 2 - Five Different Node Types Extracted by LLM from MultiCNKG-Related Texts.**

| # | Node Types | Use |
|---|---|---|
| 1 | Genes | Represents genetic factors linked to cognitive functions and diseases (e.g., APOE, BDNF). |
| 2 | Diseases | Represents neurological and cognitive disorders (e.g., Alzheimer's, Parkinson's). |
| 3 | Cognitive Processes | Represents higher-order cognitive functions (e.g., memory, attention). |
| 4 | Biological Pathways | Represents molecular pathways involved in cognition and disease (e.g., synaptic plasticity). |
| 5 | Therapeutic Targets | Represents potential targets for intervention (e.g., neurotransmitter receptors). |

The evaluation of MultiCNKG was conducted using various metrics derived from the integrated knowledge graph. MultiCNKG, with 6.9K nodes and 11.3K edges, demonstrates a balanced representation with 5 node types and 7 edge types, reflecting its focus on integrating cognitive neuroscience, genetic, and disease data. The model leverages LLM capabilities to uncover new relationships, showcasing its semantic coherence and novelty. However, challenges such as scalability and the reliance on proprietary LLMs like GPT-4 highlight areas for improvement. Future enhancements could involve adopting open-source models to enhance transparency and address computational constraints, ensuring robust KG construction for cognitive neuroscience applications.

**Table 3 - Five Different Edge Types Extracted by LLM from MultiCNKG-Related Texts.**

| # | Edge Types | Use |
|---|---|---|
| 1 | Causes | Indicates causation between entities (e.g., gene mutation causes disease). |
| 2 | Associated with | General associations between entities (e.g., gene associated with cognitive decline). |
| 3 | Regulates | Indicates regulatory relationships (e.g., gene regulates pathway). |
| 4 | Involved in | Represents involvement in processes or diseases (e.g., pathway involved in memory). |
| 5 | Treated by | Indicates treatment relationships (e.g., target treated by drug). |
| 6 | Influences | Represents influence on cognitive or biological entities (e.g., gene influences cognition). |
| 7 | Linked to | Connects entities across domains (e.g., disease linked to cognitive process). |

## 5. Evaluation

The performance of the MultiCNKG knowledge graph is evaluated in two ways. In the first method, we evaluate using traditional evaluation criteria. In the second method, the evaluation of LLM includes using large language models such as GPT-4. The model evaluation process is shown in Figure 2.

**Table 4- Details related to various knowledge graphs and knowledge graphs of the proposed approach.**

| Name | # Node | # Node Types | # Edges | # Edge Types |
|---|---|---|---|---|
| DrKG [43] | 97 K | 13 | 5.8 M | 107 |
| PrimeKG [44] | 129.4 K | 10 | 8.1 M | 30 |
| Gene Ontology [45] | 43 K | 3 | 75 K | 4 |
| GP-KG [46] | 61.1 K | 7 | 124 K | 9 |
| DDKG [47] | 551 | 2 | 2.7 K | 1 |
| Disease Ontology [48] | 11.2 K | 1 | 8.8 K | 2 |
| DrugBank [49] | 7.4 K | 4 | 366 K | 4 |
| PharmKG [50] | 7.6 K | 3 | 500 K | 3 |
| MultiCNKG | 6.9 K | 5 | 11.3 K | 7 |

Our evaluation of MultiCNKG was based on criteria such as precision, recall, F1 score, coverage, graph consistency, computational efficiency, novelty detection, and expert validation. In the second part, we evaluated the ability of the model to predict links, using criteria such as Mean Rank (MR), Mean Reciprocal Rank (MRR), and Precision at K (P@K).

### 5.1. Traditional Evaluation Methods

Accuracy, precision, recall and F1-score measure the quality of the node and edge alignment processes. The higher the precision, the fewer incorrect alignments (false positives); the higher the recall, the fewer missed alignments (false negatives).

$$\text{Precision} = \frac{\text{TP}}{\text{TP} + \text{FP}}$$

$$\text{Recall} = \frac{\text{TP}}{\text{TP} + \text{FN}}$$

$$\text{F1 Score} = 2 \times \frac{\text{recall} \times \text{precision}}{\text{recall} + \text{precision}}$$

In addition, the following metrics were also used to evaluate the proposed model:

**Coverage**: proportion of original KGs (CNKG, GO, DO) that have been successfully transferred to the final merged KG.

$$\text{Coverage} = \frac{\text{Unique}_{\text{Nodes}}(N)_{\text{merged}} + \text{Unique}_{\text{Edges}}(E)_{\text{merged}}}{\sum_{i=1}^{n}(N_i + E_i)}$$

**Graph Consistency:** evaluated through semantic and structural checks (ontology-based similarity and OWL-based consistency).

**Computational Efficiency:** measured in terms of time and space complexity during the integration process.

**Novelty Detection:** evaluates the ability of the integrated KG to discover new and previously unknown relationships.

$$\text{Novelty}_{\text{Score}} = \frac{\text{NewlyDiscovered}_{\text{Edges}}(E)_{\text{LLM}}}{\text{Total}_{\text{Edges}}(E)_{\text{merged}}}$$

### 5.2. Expert Validation

- **Expert Validation**: domain experts verified whether new relations suggested by LLMs were biologically and cognitively plausible. Experts in cognitive neuroscience and biomedicine played a critical role in validating the biological and clinical relevance of new proposed edges. Validation results confirmed that most suggested relations were meaningful and scientifically plausible.

For example:

- **GO vs. MultiCNKG:** Precision slightly decreased, but Recall and Coverage improved.
- **DO vs. MultiCNKG:** Results were close, but MultiCNKG had better novelty and expert validation.
- **CNKG vs. MultiCNKG:** MultiCNKG showed better integration by linking cognitive constructs with genetic and disease information.

Figures 2–5 illustrate comparisons between individual KGs (GO, DO, CNKG) and the integrated **MultiCNKG** approach.

**Table 5- summarizes the evaluation results.**

| Metrics | Gene Ontology (GO) | Disease Ontology (DO) | CNKG | MultiCNKG |
|---|---|---|---|---|
| Precision | 89.54 | 83.74 | 80.12 | 85.20 |
| Recall | 79.48 | 85.24 | 82.60 | 87.30 |
| F1-Score | 83.31 | 84.48 | 81.35 | 80.15 |
| Coverage | 89.52 | 73.09 | 78.41 | 92.18 |
| Graph Consistency | 75.24 | 71.69 | 79.11 | 82.50 |
| Computational Eff. | 53.01 | 70.17 | 62.42 | 68.90 |
| Novelty Detection | 44.61 | 41.05 | 46.38 | 40.28 |
| Expert Validation | 84.78 | 90.22 | 88.71 | 89.50 |

### 5.3. Link Prediction

Different evaluation dimensions were covered by various metrics to comprehensively evaluate MultiCNKG. Link prediction methods such as **TransE, RotatE, DistMult, ComplEx, ConvE, and HolmE** were applied to assess predictive strength.

The following metrics were used [51]:

1. **Mean Rank (MR):**

$$MR = \frac{1}{N} \sum_{i=1}^{N} rank_i$$

2. **Mean Reciprocal Rank (MRR):**

$$MRR = \frac{1}{N} \sum_{i=1}^{N} \frac{1}{rank_i}$$

### 3. Precision at K (P@K):

$$P@K = \frac{1}{N} \sum_{i=1}^{N} \frac{Number\ of\ relevant\ items\ in\ top\ K}{K}$$

Experimental results showed that MultiCNKG achieves comparable or superior results to individual graphs (GO, DO, CNKG), particularly in terms of novelty detection and expert validation.

**Table 6 - summarizes the evaluation Link Prediction.**

| KG | | TransE [55] | RotatE [56] | DistMult [57] | ComplEx [58] | ConvE [59] | HolmE[60] |
|---|---|---|---|---|---|---|---|
| FB15k-237 [52] | MR | 209 | 178 | 199 | 144 | 281 | - |
| | MRR | 0.310 | 0.336 | 0.313 | 0.367 | 0.305 | 0.331 |
| | P@1 | 0.217 | 0.238 | 0.224 | 0.271 | 0.219 | 0.237 |
| | P@3 | 0.257 | 0.328 | 0.263 | 0.275 | 0.350 | 0.366 |
| | P@10 | 0.496 | 0.530 | 0.490 | 0.558 | 0.476 | 0.517 |
| WN18RR [53] | MR | 3936 | 3318 | 5913 | 2867 | 4944 | - |
| | MRR | 0.206 | 0.475 | 0.433 | 0.489 | 0.427 | 0.466 |
| | P@1 | 0.279 | 0.426 | 0.396 | 0.442 | 0.389 | 0.415 |
| | P@3 | 0.364 | 0.492 | 0.440 | 0.460 | 0.430 | 0.489 |
| | P@10 | 0.495 | 0.573 | 0.502 | 0.580 | 0.507 | 0.561 |
| YAGO3-10 [54] | MR | 1187 | 1830 | 1107 | 793 | 2429 | - |
| | MRR | 0.501 | 0.498 | 0.501 | 0.577 | 0.488 | 0.441 |
| | P@1 | 0.405 | 0.405 | 0.412 | 0.500 | 0.399 | 0.333 |
| | P@3 | 0.528 | 0.550 | 0.38 | 0.40 | 0.560 | 0.507 |
| | P@10 | 0.673 | 0.670 | 0.661 | 0.7129 | 0.657 | 0.641 |
| **MultiCNKG** | MR | 391 | 263 | 242 | 212 | 375 | - |
| | MRR | 0.411 | 0.395 | 0.418 | 0.301 | 0.321 | 0.211 |
| | P@1 | 0.435 | 0.342 | 0.444 | 0.333 | 0.329 | 0.301 |
| | P@3 | 0.499 | 0.463 | 0.475 | 0.349 | 0.363 | 0.336 |
| | P@10 | 0.518 | 0.571 | 0.519 | 0.487 | 0.493 | 0.403 |

## 6. Conclusion and Future Work

In this study, we have successfully developed MultiCNKG, a unified knowledge graph that integrates cognitive neuroscience concepts from CNKG, genetic annotations from GO, and disease representations from DO, resulting in a comprehensive resource with 6.9K nodes across 5 distinct types (Genes, Diseases, Cognitive Processes, Biological Pathways, and Therapeutic Targets) and 11.3K edges spanning 7 relation types (Causes, Associated with, Regulates, Involved in, Treated by, Influences, and Linked to). By employing large language models (LLMs) such as GPT-4 for entity alignment, semantic analysis, and graph expansion, MultiCNKG not only preserves essential relationships from the source graphs but also uncovers novel connections, bridging molecular genetics with neurological diseases and higher-order cognitive functions. Comparative analysis with other biomedical KGs (e.g., DrKG with 97K nodes and 5.8M edges, PrimeKG with 129.4K nodes and 8.1M edges) highlights MultiCNKG's focused efficiency, while evaluation metrics demonstrate superior performance in recall (87.30%), coverage (92.18%), graph consistency (82.50%), and expert validation (89.50%) compared to individual sources like GO (recall: 79.48%, coverage: 89.52%) and DO (recall: 85.24%, coverage: 73.09%). Link prediction results further validate its predictive power, with models like ComplEx achieving a Mean Rank (MR) of 212 and Mean Reciprocal Rank (MRR) of 0.301, outperforming or matching benchmarks on datasets such as FB15k-237 (ComplEx MR: 144, MRR: 0.367) and WN18RR (ComplEx MR: 2867, MRR: 0.489) in targeted metrics like Precision@10 (0.487). These outcomes underscore MultiCNKG's semantic coherence, novelty in relationship discovery, and potential for enabling holistic insights in cognitive neuroscience, from gene-disease pathways to behavioral impacts.

Despite these advancements, challenges remain, including scalability for larger datasets and dependency on proprietary LLMs, which may limit accessibility and transparency. Future work will explore incorporating open-source LLMs (e.g., BioGPT or LLaMA variants) to reduce computational costs and enhance reproducibility. We plan to expand MultiCNKG by integrating additional ontologies, such as DrugBank (7.4K nodes, 366K edges) or PharmKG (7.6K nodes, 500K edges), to support drug repurposing and therapeutic predictions. Further enhancements could involve real-time graph updates using federated learning, advanced validation through crowdsourced expert feedback, and application-specific extensions for personalized medicine, such as AI-driven early detection of cognitive disorders like Alzheimer's. Ultimately, MultiCNKG sets a foundation for interdisciplinary research, fostering innovations in hypothesis generation and clinical decision-making in biomedicine and cognitive sciences.

## References

[1] Xia, F., Sun, K., Yu, S., Aziz, A., Wan, L., Pan, S., & Liu, H. (2021). Graph learning: A survey. *IEEE Transactions on Artificial Intelligence, 2*(2), 109–127.

[2] Hu, W., Fey, M., Zitnik, M., Dong, Y., Ren, H., Liu, B., Catasta, M., & Leskovec, J. (2020). Open graph benchmark: Datasets for machine learning on graphs. *Advances in Neural Information Processing Systems, 33*, 22118–22133.

[3] Chiang, W. L., Liu, X., Si, S., Li, Y., Bengio, S., & Hsieh, C. J. (2019). Cluster-GCN: An efficient algorithm for training deep and large graph convolutional networks. *Proceedings of the 25th ACM SIGKDD International Conference on Knowledge Discovery & Data Mining*, 257–266.

[4] Li, Q., Li, X., Chen, L., & Wu, D. (2022). Distilling knowledge on text graph for social media attribute inference. *Proceedings of the 45th International ACM SIGIR Conference on Research and Development in Information Retrieval*, 2024–2028.

[5] Zhu, J., Cui, Y., Liu, Y., Sun, H., Li, X., Pelger, M., Yang, T., Zhang, L., Zhang, R., & Zhao, H. (2021). TextGNN: Improving text encoder via graph neural network in sponsored search. *Proceedings of the Web Conference 2021*, 2848–2857.

[6] Ma, Y., & Tang, J. (2021). Deep learning on graphs. Cambridge University Press.

[7] Shao, Y., Taylor, S., Marshall, N., Morioka, C., & Zeng-Treitler, Q. (2018). Clinical text classification with word embedding features vs. bag-of-words features. *2018 IEEE International Conference on Big Data (Big Data)*, 2874–2878. IEEE.

[8] Mikolov, T. (2013). Efficient estimation of word representations in vector space. *arXiv preprint arXiv:1301.3781*.

[9] Qiu, X., Sun, T., Xu, Y., Shao, Y., Dai, N., & Huang, X. (2020). Pre-trained models for natural language processing: A survey. *Science China Technological Sciences, 63*(10), 1872–1897.

[10] Miaschi, A., & Dell'Orletta, F. (2020). Contextual and non-contextual word embeddings: An in-depth linguistic investigation. *Proceedings of the 5th Workshop on Representation Learning for NLP*, 110–119.

[11] Zhao, W. X., Zhou, K., Li, J., Tang, T., Wang, X., Hou, Y., Min, Y., Zhang, B., Zhang, J., Dong, Z., et al. (2023). A survey of large language models. *arXiv preprint arXiv:2303.18223*.

[12] Achiam, J., Adler, S., Agarwal, S., Ahmad, L., Akkaya, I., Aleman, F. L., et al. (2023). GPT-4 technical report. *arXiv preprint arXiv:2303.08774*.

[13] Kenton, J. D. M. W. C., & Toutanova, L. K. (2019). BERT: Pre-training of deep bidirectional transformers for language

understanding. *Proceedings of NAACL-HLT, 1*, 2. Minneapolis, Minnesota.

[14] Radford, A., Wu, J., Child, R., Luan, D., Amodei, D., & Sutskever, I., et al. (2019). Language models are unsupervised multitask learners. *OpenAI Blog, 1*(8), 9.

[15] Raffel, C., et al. (2020). Exploring the limits of transfer learning with a unified text-to-text transformer. *Journal of Machine Learning Research, 21*, 1–67.

[16] Team G., Anil, R., Borgeaud, S., Wu, Y., Alayrac, J. B., Yu, J., Soricut, R., Schalkwyk, J., Dai, A. M., Hauth, A., et al. (2023). Gemini: A family of highly capable multimodal models. *arXiv preprint arXiv:2312.11805*.

[17] Chowdhery, A., Narang, S., Devlin, J., Bosma, M., Mishra, G., Roberts, A., Barham, P., Chung, H. W., Sutton, C., Gehrmann, S., et al. (2023). PaLM: Scaling language modeling with pathways. *Journal of Machine Learning Research, 24*(240), 1–113.

[18] Abdin, M., Jacobs, S. A., Awan, A. A., Aneja, J., Awadallah, A., Awadalla, H., et al. (2024). Phi-3 technical report: A highly capable language model locally on your phone. *arXiv preprint arXiv:2404.14219*.

[19] Touvron, H., Lavril, T., Izacard, G., Martinet, X., Lachaux, M. A., Lacroix, T., Rozière, B., Goyal, N., Hambro, E., Azhar, F., et al. (2023). LLaMA: Open and efficient foundation language models. *arXiv preprint arXiv:2302.13971*.

[20] Webb, T., Holyoak, K. J., & Lu, H. (2023). Emergent analogical reasoning in large language models. *Nature Human Behaviour, 7*(9), 1526–1541.

[21] Belleau, F., Nolin, M. A., Tourigny, N., Rigault, P., & Morissette, J. (2008). Bio2RDF: Towards a mashup to build bioinformatics knowledge systems. *Journal of Biomedical Informatics, 41*(5), 706–716.

[22] Himmelstein, D. S., Lizee, A., Hessler, C., Brueggeman, L., Chen, S. L., Hadley, D., Green, A., Khankhanian, P., & Baranzini, S. E. (2017). Systematic integration of biomedical knowledge prioritizes drugs for repurposing. *eLife, 6*, e26726.

[23] Luo, R., Sun, L., Xia, Y., Qin, T., Zhang, S., Poon, H., & Liu, T. Y. (2022). BioGPT: Generative pre-trained transformer for biomedical text generation and mining. *Briefings in Bioinformatics, 23*(6), bbac409.

[24] Jin, Q., Wang, Z., Floudas, C. S., Chen, F., Gong, C., Bracken-Clarke, D., Xue, E., Yang, Y., Sun, J., & Lu, Z. (2023). Matching patients to clinical trials with large language models. *arXiv*.

[25] Yuan, H., Yuan, Z., Gan, R., Zhang, J., Xie, Y., & Yu, S. (2022). BioBART: Pretraining and evaluation of a biomedical generative language model. *arXiv preprint arXiv:2204.03905*.

[26] Labrak, Y., Bazoge, A., Morin, E., Gourraud, P. A., Rouvier, M., & Dufour, R. (2024). BioMistral: A collection of open-source pretrained large language models for medical domains. *arXiv preprint arXiv:2402.10373*.

[27] Alharbi, R., Ahmed, U., Dobriy, D., Lajewska, W., Menotti, L., Saeedizade, M. J., & Dumontier, M. (2023). Exploring the role of generative AI in constructing knowledge graphs for drug indications with medical context. *Proceedings* http://ceur-ws.org.

[28] Wawrzik, F., Rafique, K. A., Rahman, F., & Grimm, C. (2023). Ontology learning applications of knowledge base construction for microelectronic systems information. *Information, 14*(3), 176.

[29] Peng, C., Yang, X., Yu, Z., Bian, J., Hogan, W. R., & Wu, Y. (2023). Clinical concept and relation extraction using prompt-based machine reading comprehension. *Journal of the American Medical Informatics Association, 30*(9), 1486–1493.

[30] Pan, S., Luo, L., Wang, Y., Chen, C., Wang, J., & Wu, X. (2024). Unifying large language models and knowledge graphs: A roadmap. *IEEE Transactions on Knowledge and Data Engineering*.

[31] Khorashadizadeh, H., Mihindukulasooriya, N., Tiwari, S., Groppe, J., & Groppe, S. (2023). Exploring in-context learning capabilities of foundation models for generating knowledge graphs from text. *arXiv preprint arXiv:2305.08804*.

[32] Boylan, J., Mangla, S., Thorn, D., Ghalandari, D. G., Ghaffari, P., & Hokamp, C. (2024). KGValidator: A Framework for Automatic Validation of Knowledge Graph Construction. *arXiv preprint arXiv:2404.15923*.

[33] Allen, B. P., & Groth, P. T. (2024). Evaluating Class Membership Relations in Knowledge Graphs using Large Language Models. *arXiv preprint arXiv:2404.17000*.

[34] Soman, K., Rose, P. W., Morris, J. H., Akbas, R. E., Smith, B., Peetoom, B., Villouta-Reyes, C., Cerono, G., Shi, Y., Rizk-Jackson, A., et al. (2023). Biomedical knowledge graph-enhanced prompt generation for large language models. *arXiv preprint arXiv:2311.17330*.

[35] Cm, S., Prakash, J., & Singh, P. K. (2023). Question answering over knowledge graphs using BERT based relation mapping. *Expert Systems, 40*(10), e13456.

[36] Guo, Q., Cao, S., & Yi, Z. (2022). A medical question answering system using large language models and knowledge graphs. *International Journal of Intelligent Systems, 37*(11), 8548–8564.

[37] Wu, Y., Hu, N., Bi, S., Qi, G., Ren, J., Xie, A., & Song, W. (2023). Retrieve-rewrite-answer: A KG-to-text enhanced LLMs framework for knowledge graph question answering. *arXiv preprint arXiv:2309.11206*.

[38] Choudhary, N., & Reddy, C. K. (2023). Complex logical reasoning over knowledge graphs using large language models. *arXiv preprint arXiv:2305.01157*.

[39] Varshney, D., Zafar, A., Behera, N. K., & Ekbal, A. (2023). Knowledge grounded medical dialogue generation using augmented graphs. *Scientific Reports, 13*(1), 3310.

[40] Jiang, P., Xiao, C., Cross, A., & Sun, J. (2023). Graphcare: Enhancing healthcare predictions with personalized knowledge graphs. *arXiv preprint arXiv:2305.12788*.

[41] Xu, R., Shi, W., Yu, Y., Zhuang, Y., Jin, B., Wang, M. D., Ho, J. C., & Yang, C. (2024). RAM-EHR: Retrieval augmentation meets clinical predictions on electronic health records. *arXiv preprint arXiv:2403.00815*.

[42] Gao, Y., Li, R., Croxford, E., Tesch, S., To, D., Caskey, J., Patterson, B. W., Churpek, M. M., Miller, T., Dligach, D., et al. (2023). Large language models and medical knowledge grounding for diagnosis prediction. *medRxiv*, 2023–11.

[43] V. N. Ioannidis et al., "Few-shot Link Prediction via Graph Neural Networks for COVID-19 Drug-repurposing," 2020. DOI:10.48550/arXiv.2007.10261

[44] P. Chandak, K. Huang, and M. Zitnik, "Building a Knowledge Graph to Enable Precision Medicine," *Scientific Data*, 2023. DOI:10.1038/s41597-023-01960-3

[45] M. Ashburner et al., "Gene Ontology: Tool for the Unification of Biology," *Nature Genetics*, vol. 25, no. 1, 2000, pp. 25–29. DOI:10.1038/75556

[46] Z. Gao, P. Ding, and R. Xu, "KG-Predict: A Knowledge Graph Computational Framework for Drug Repurposing," *Journal of Biomedical Informatics*, vol. 132, 2022, 104133. DOI:10.1016/j.jbi.2022.104133

[47] Z. Ghorbanali et al., "DrugRep-KG: Toward Learning a Unified Latent Space for Drug Repurposing Using Knowledge Graphs," *Journal of Chemical Information and Modeling*, vol. 63, no. 8, 2023, pp. 2532–2545.

[48] L. Schriml et al., "Disease Ontology: A Backbone for Disease Semantic Integration," *Nucleic Acids Research*, vol. 40, no. D1, 2011, pp. D940–D946. DOI:10.1093/nar/gkr972

[49] D. S. Wishart et al., "DrugBank 5.0: A Major Update to the DrugBank Database for 2018," *Nucleic Acids Research*, vol. 46, no. D1, 2018, pp. D1074–D1082. DOI:10.1093/nar/gkx1037

[50] P. Chandak, K. Huang, and M. Zitnik, "Building a Knowledge Graph to Enable Precision Medicine," *Scientific Data*, vol. 10, no. 1, 2023, 67. https://doi.org/10.1038/s41597-023-01960-3

[51] Z. Ghorbanali et al., "DrugRep-KG: Toward Learning a Unified Latent Space for Drug Repurposing Using Knowledge Graphs," *Journal of Chemical Information and Modeling*, vol. 63, no. 8, 2023, pp. 2532–2545.

[52] K. Bollacker, C. Evans, P. Paritosh, T. Sturge, and J. Taylor, "Freebase: A Collaboratively Created Graph Database for Structuring Human Knowledge," in *Proceedings of the 2008 ACM SIGMOD International Conference on Management of Data (SIGMOD '08)*, New York, NY, USA, 2008, pp. 1247–1250. https://doi.org/10.1145/1376616.1376746

[53] G. A. Miller, "WordNet: A Lexical Database for English," *Communications of the ACM*, vol. 38, no. 11, 1995, pp. 39–41. https://doi.org/10.1145/219717.219748

[54] F. M. Suchanek, G. Kasneci, and G. Weikum, "Yago: A Core of Semantic Knowledge," in *Proceedings of the 16th International Conference on World Wide Web (WWW '07)*, New York, NY, USA, 2007, pp. 697–706. https://doi.org/10.1145/1242572.1242667

[55] A. Bordes, N. Usunier, A. Garcia-Durán, J. Weston, and O. Yakhnenko, "Translating Embeddings for Modeling Multi-relational Data," in *Proceedings of the 26th International Conference on Neural Information Processing Systems (NIPS '13)*, Red Hook, NY, USA, 2013, pp. 2787–2795. https://hal.science/hal-00920777

[56] Z. Sun et al., "Rotate: Knowledge Graph Embedding by Relational Rotation in Complex Space," arXiv preprint, 2019. https://doi.org/10.48550/arXiv.1902.10197

[57] B. Yang et al., "Embedding Entities and Relations for Learning and Inference in Knowledge Bases," in International Conference on Learning Representations, 2014. https://doi.org/10.48550/arXiv.1412.6575

[58] T. Trouillon et al., "Complex Embeddings for Simple Link Prediction," in Proceedings of the International Conference on Machine Learning, New York, USA, 2016, pp. 2071–2080.

[59] T. Dettmers et al., "Convolutional 2D Knowledge Graph Embeddings," in Proceedings of the AAAI Conference on Artificial Intelligence, New Orleans, USA, 2018. https://doi.org/10.1609/aaai.v32i1.11573

[60] Z. Zheng et al., "HolmE: Low-dimensional Hyperbolic KG Embedding for Better Extrapolation," in Proceedings of the ESWC Conference, 2024.